\documentclass[sigconf]{acmart}

\usepackage{bm}
\usepackage{amsmath} 
\usepackage{subcaption}
\usepackage{algorithm}
\usepackage{algorithmic}
\usepackage{multirow} 
\usepackage{tabularx} 
\usepackage{pifont}
\usepackage{booktabs}

\newcommand{\cmark}{\ding{51}}
\newcommand{\xmark}{\ding{55}}

\AtBeginDocument{%
  }

\setcopyright{acmlicensed}
\copyrightyear{2025}
\acmYear{2025}
\acmDOI{XXXXXXX.XXXXXXX}
\acmConference[Conference acronym 'XX]{Make sure to enter the correct
  conference title from your rights confirmation email}{June 03--05,
  2018}{Woodstock, NY}
\acmISBN{978-1-4503-XXXX-X/2018/06}

\begin{document}

\title{RePainter: Empowering E-commerce Object Removal via Spatial-matting Reinforcement Learning}

\settopmatter{authorsperrow=4}

\author{Zipeng Guo}
\authornote{Equal contribution.}
\affiliation{
  \institution{Sun Yat-sen University}
  \city{Shenzhen}
  \country{China}}
\email{guozp5@mail2.sysu.edu.cn}

\author{Lichen Ma}
\authornotemark[1]
\authornotemark[2]
\affiliation{%
  \institution{JD.COM}
  \city{Beijing}
  \country{China}
}
\email{malichen2020@gmail.com}

\author{Xiaolong Fu}
\authornotemark[1]
\affiliation{%
  \institution{JD.COM}
  \city{Beijing}
  \country{China}
}
\email{fxlcumt@gmail.com}

\author{Gaojing Zhou}
\affiliation{%
  \institution{JD.COM}
  \city{Beijing}
  \country{China}
}
\email{darkflameofmaster@gmail.com}

\author{Lan Yang}
\affiliation{%
  \institution{Beijing University of Chemical Technology}
  \city{Beijing}
  \country{China}
}
\email{2024400283@buct.edu.cn}

\author{Yuchen Zhou}
\affiliation{
  \institution{Sun Yat-sen University}
  \city{Shenzhen}
  \country{China}}
\email{zhouych37@mail2.sysu.edu.cn}

\author{Linkai Liu}
\affiliation{%
  \institution{Sun Yat-sen University}
  \city{Shenzhen}
  \country{China}
}
\email{liulk6@mail2.sysu.edu.cn}

\author{Yu He}
\affiliation{%
  \institution{JD.COM}
  \city{Beijing}
  \country{China}
}
\email{heyu2579@gmail.com}

\author{Ximan Liu}
\affiliation{%
  \institution{JD.COM}
  \city{Beijing}
  \country{China}
}
\email{liuximan.3@jd.com}

\author{Shiping Dong}
\affiliation{%
  \institution{Hunan University}
  \city{Hunan}
  \country{China}
}
\email{dongshiping@hnu.edu.cn}

\author{Jingling Fu}
\affiliation{%
  \institution{JD.COM}
  \city{Beijing}
  \country{China}
}
\email{fjlzzf@gmail.com}

\author{Zhen Chen}
\affiliation{%
  \institution{JD.COM}
  \city{Beijing}
  \country{China}
}
\email{chenzhen48@jd.com}

\author{Yu Shi}
\affiliation{%
  \institution{JD.COM}
  \city{Beijing}
  \country{China}
}
\email{37675890@qq.com}

\author{Junshi Huang}
\authornote{Project leader.}
\affiliation{%
  \institution{JD.COM}
  \city{Beijing}
  \country{China}
}
\email{junshi.huang@gmail.com}

\author{Jason Li}
\affiliation{%
  \institution{JD.COM}
  \city{Beijing}
  \country{China}
}
\email{lixiumei.40@jd.com}

\author{Chao Gou}
\authornote{Corresponding author.}
\affiliation{%
  \institution{Sun Yat-sen University}
  \city{Shenzhen}
  \country{China}
}
\email{gouchao@mail.sysu.edu.cn}

\renewcommand{\shortauthors}{Zipeng Guo, et al.}

\renewcommand{\shortauthors}{Trovato et al.}

\begin{abstract}
In web data, product images are central to boosting user engagement and advertising efficacy on e-commerce platforms, yet the intrusive elements such as watermarks and promotional text remain major obstacles to delivering clear and appealing product visuals. Although diffusion-based inpainting methods have advanced, they still face challenges in commercial settings due to unreliable object removal and limited domain-specific adaptation. To tackle these challenges, we propose Repainter, a reinforcement learning framework that integrates spatial-matting trajectory refinement with Group Relative Policy Optimization (GRPO). Our approach modulates attention mechanisms to emphasize background context, generating higher-reward samples and reducing unwanted object insertion. We also introduce a composite reward mechanism that balances global, local, and semantic constraints, effectively reducing visual artifacts and reward hacking. Additionally, we contribute EcomPaint-100K, a high-quality, large-scale e-commerce inpainting dataset, and a standardized benchmark EcomPaint-Bench for fair evaluation. Extensive experiments demonstrate that Repainter significantly outperforms state-of-the-art methods, especially in challenging scenes with intricate compositions. We will release our code and weights upon acceptance. 
\end{abstract}




\maketitle

\begin{figure*}[t]
  \includegraphics[width=\textwidth]{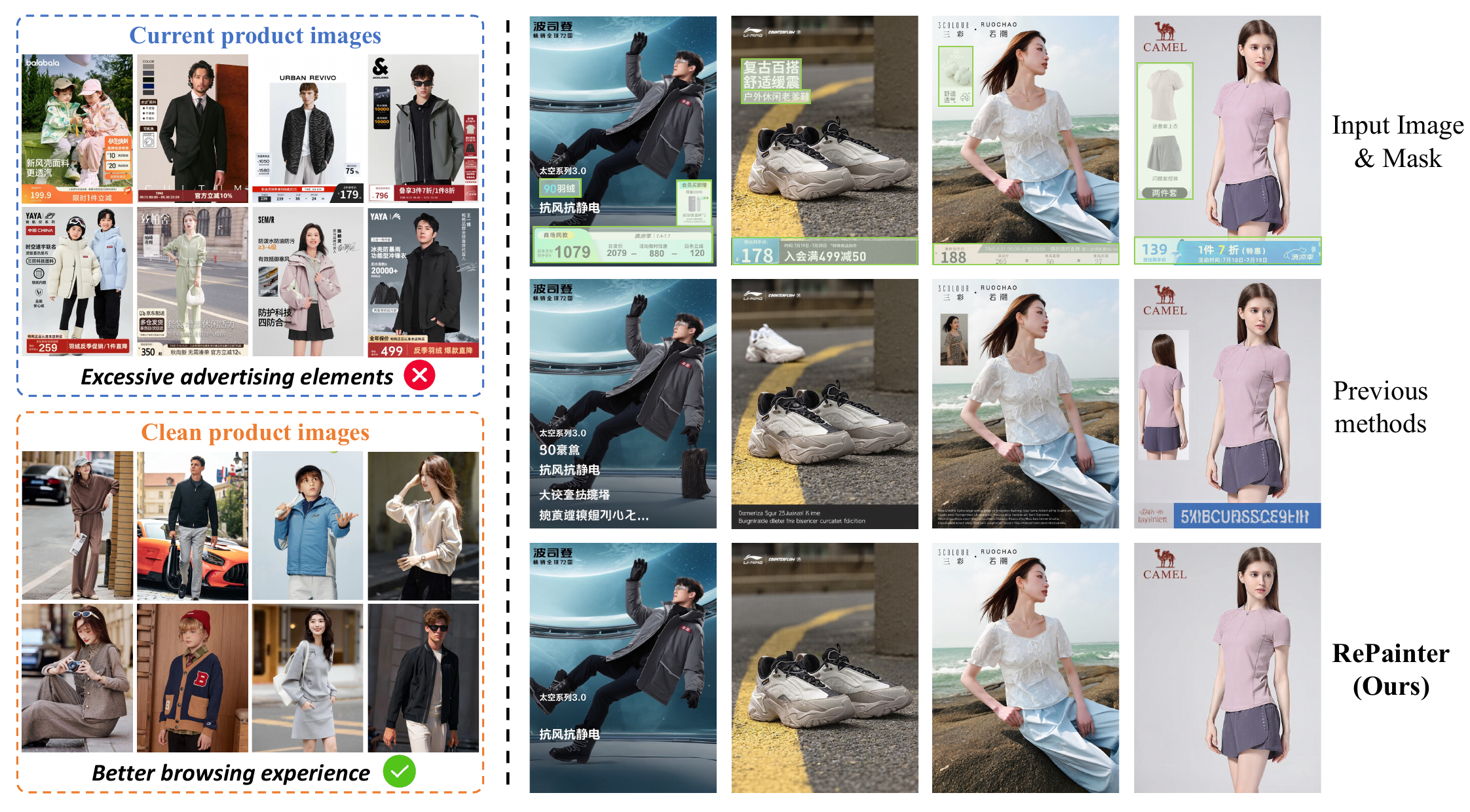}
  \caption{(Left) Excessive advertising elements in product images (e.g., price tags and text) often compromise the visual appeal of the images and adversely impact the user browsing experience. (Right) Previous methods~\cite{FluxControlnetInpainting2024,gong2025onereward,BlackForestLabsFlux2024,wang2025towards} tend to generate unintentional objects and struggle to remove the target object’s effects, leading to unrealistic outputs. In contrast, our RePainter seamlessly removes target objects while ensuring visual coherence in the generated images.}
  \label{fig:teaser}
\end{figure*}

\section{Introduction}
In e-commerce scenarios, product images play a pivotal role in attracting user attention and boosting advertising efficacy~\cite{mishra2020learning,ku2023staging,zhao2025dreampainter}. However, most product images contain excessive advertising elements, such as watermarks, price tags, and promotional text. These visual distractions substantially compromise product clarity and visual appeal, leading to diminished user browsing experience and negatively impacting purchase intent, as shown in Figure~\ref{fig:teaser}. Consequently, there is an increasing demand for automated solutions that can accurately remove such elements to deliver a more immersive shopping experience for users.

Recent advancements in diffusion models~\cite{ho2020denoising,Podell2023SDXL,rombach2022high,Esser2024Scaling} have greatly improved image generation~\cite{BlackForestLabsFlux2024,Gao2025Seedream,Zheng2024Cogview3} and editing~\cite{Brooks2023InstructPix2Pix,Deng2025EmergingProperties,Ge2024SeedX,liu2025step1x-edit,wu2025omnigen2} capabilities. Among these, image inpainting~\cite{Ju2024BrushNet,Lugmayr2022Repaint,Suvorov2021FourierInpainting,Zhuang2024TaskPrompt,yu2025omnipaint,FluxControlnetInpainting2024}, aiming to reconstruct coherent content in masked regions while preserving unmasked pixels, has emerged as a key technique for object removal, offering a promising solution to the challenges outlined above. Despite these advancements, the performance of existing inpainting frameworks~\cite{FluxControlnetInpainting2024,gong2025onereward,BlackForestLabsFlux2024,wang2025towards,wei2025omnieraser,Liu2025EraseDiffusion,Sun2025AttentiveEraser,Jiang2025Smarteraser} still remains suboptimal in e-commerce scenarios, hindered by two primary limitations. First, these approaches frequently fail to ensure reliable object removal, often introducing artifacts, hallucinated objects, or inconsistent textures that disrupt visual coherence. Second, most methods rely on supervised fine-tuning (SFT) trained on generic datasets that lack of e-commerce-specific imagery, resulting in backgrounds that suffer from poor resolution or mismatched stylistic elements when applied to product images. While SFT provides a direct approach to adapt pre-trained models, its effectiveness is highly dependent on access to large-scale, high-quality, and domain-specific paired datasets. The scarcity of such specialized data severely limits the generalization capability of SFT-based models, causing them to often overfit to the narrow scenarios presenting in the limited fine-tuning data.

Reinforcement learning (RL) based framework~\cite{Sutton1998RL,Schulman2017PPO,Wallace2024DMAlignment,Guo2025CoTImageGen,Black2023RL4Diffusion,Fan2023DPOK} has emerged as a promising solution to address these issues. Specifically, methods~\cite{liu2025flowgrpo,Xue2025DanceGRPO} based on Group Relative Policy Optimization (GRPO)~\cite{Guo2025DeepSeekR1,Shao2024DeepSeekMath}, have recently been studied, achieving optimal alignment with domain-specific visual preferences. Although effective, current GRPO-based approaches~\cite{liu2025flowgrpo,Xue2025DanceGRPO,Li2025MixGRPO} are constrained by inherent limitations in stochastic exploration, resulting in similar reward scores for comparable images within the same group, and a limited capacity to explore high-reward samples within the action space. This inefficient exploration leads to slow convergence and unstable training, potentially trapping the model in suboptimal local solutions and ultimately hindering RL from realizing its full potential in complex image inpainting.  

To address these challenges, we propose Repainter, a novel reinforcement learning framework that integrates spatial-matting trajectory refinement with GRPO for high-fidelity e-commerce image inpainting. Our method enhances the sampling trajectory by modulating spatial attention mechanisms to prioritize background context over distracting foreground elements, thereby reducing unwanted object insertion and improving semantic coherence. During the roll-out phase, this strategy increases the likelihood of positive rewards during exploration, significantly accelerating convergence while improving overall performance. Furthermore, we introduce a local-global composite reward mechanism that jointly optimizes global structural consistency, local reconstruction accuracy, and semantic validity. This integrated approach provides a more comprehensive optimization signal and mitigates the risk of reward hacking, where models over-optimize a single metric at the expense of overall image quality. To bridge the scarcity of high-quality data for e-commerce inpainting tasks, we build a high-quality large-scale dataset, EcomPaint-100K, specifically curated for e-commerce image manipulation, which captures diverse product categories, backgrounds, and professional photography standards. Additionally, we introduce the EcomPaint-Bench to provide a robust and comprehensive evaluation platform for model performance.
 
Extensive experiments demonstrate that Repainter outperforms state-of-the-art methods across various key metrics. Our method effectively suppresses common artifacts such as text hallucination and color inconsistency, offering a robust solution for generating clean, professional product images that improve user experience and meet e-commerce requirements. To the best of our knowledge, this is the first work that utilizes GRPO for image inpainting. 

We summarize our contributions as four-folds:
\begin{itemize}
\item We propose Repainter, a novel GRPO-based inpainting framework that optimize sampling trajectories via spatial-matting guidance to mitigate object hallucination.
\item We design a composite local-global reward mechanism that jointly optimizes global structure, local reconstruction, and semantic validity, effectively mitigating reward hacking.
\item We introduce EcomPaint-100K, a large-scale high-quality dataset for e-commerce inpainting, along with the EcomPaint-Bench for standardized evaluation.
\item Extensive experiments and user studies demonstrate the effectiveness of our method, with both removal quality and stability surpassing SOTA methods.
\end{itemize}

\section{Related Works}
\noindent \textbf{Image inpainting.}
Image inpainting, a critical technique in image generation, focuses on restoring missing image regions by leveraging existing contextual information, with broad applications in object removal, insertion, replacement, and background generation. With the advent of deep learning, methods~\cite{Cao2022PriorFeature,Li2022MAT,Pathak2016ContextEncoders,Yu2019FreeForm,Zhao2020CoModGAN} based on Generative Adversarial Networks (GAN)~\cite{Goodfellow2014GAN} became dominant. Notably, LaMa~\cite{suvorov2021resolution} introduced Fast Fourier Convolutions, significantly improving the ability to handle large and complex masks while preserving global structural consistency. More recently, approaches~\cite{Lugmayr2022Repaint,rombach2022high,Avrahami2023BlendedLatentDiffusion,Xie2023SmartBrush,Zhang2023MagicBrush,Yildirim2023InstInpaint,BlackForestLabsFlux2024,FluxControlnetInpainting2024} based on diffusion models have attracted significant attention due to their superior generative quality. Among them, RePaint~\cite{Lugmayr2022Repaint} was an early method that applied a pre-trained unconditional diffusion model to inpainting by repeatedly sampling the unknown region and blending it with the known context. Subsequent models, such as SD-Inpaint~\cite{rombach2022high} Blended Latent Diffusion~\cite{Avrahami2023BlendedLatentDiffusion} and SmartBrush~\cite{Xie2023SmartBrush}, adopted a more efficient approach by concatenating the latent representations of the mask and the source image as input to diffusion models. This paradigm established a strong foundation for high-fidelity, text-guided editing. Follow-up works, such as MagicBrush~\cite{Zhang2023MagicBrush} and Inst-Inpaint~\cite{Yildirim2023InstInpaint}, introduced more refined instruction-based datasets to improve the accuracy of image editing. Recently, FLUX Fill~\cite{BlackForestLabsFlux2024} has emerged as a powerful baseline demonstrating strong performance in both inpainting and outpainting. In the e-commerce domain, where the goal is to generate visually appealing backgrounds for product images, traditional inpainting methods often fall short due to their lack of domain-specific knowledge. While some recent works~\cite{Alimama2024EcomXL,FluxControlnetInpainting2024} have focused on e-commerce scenarios, their dependence on low-aesthetic training data limits the quality of generated results. To bridge these gaps, we construct the EcomPaint-100K dataset, comprising 100K high-quality e-commerce images for training. Furthermore, we introduce RePainter, a robust inpainting framework specifically designed to meet the demanding requirements of product image generation.

\noindent \textbf{Erase inpainting.}
Erase inpainting, one specialized form of image inpainting, focuses on removing unwanted content from images using input masks or text instructions. Text-driven approaches~\cite{Fu2024GuidingInstruction,Yildirim2023InstInpaint,Yu2024PromptFix,liu2025step1x-edit,wu2025omnigen2} specify objects for removal via instructions but are constrained by text embedding performance~\cite{MarcosManchon2024OpenVocabulary,Yang2024PhraseGrounding}, particularly in handling multiple objects and attribute understanding. In contrast, mask-guided methods~\cite{wei2025omnieraser,yu2025omnipaint,Li2024MagicEraser,Zhuang2025TaskPrompt,Ekin2024ClipAway,Sun2025AttentiveEraser,Jiang2025Smarteraser,Liu2025EraseDiffusion,FluxControlnetInpainting2024,gong2025onereward,BlackForestLabsFlux2024,wang2025towards} provide more precise control. Recent advances, such as MagicEraser~\cite{Li2024MagicEraser}, generate removal data by shifting objects within an image, while SmartEraser~\cite{Jiang2025Smarteraser} synthesizes a million-sample dataset using alpha blending. ASUKA~\cite{wang2025towards} enhances image inpainting with color-consistency and mitigate object hallucination while leveraging the generation capacity of the frozen inpainting model. OmniPaint~\cite{yu2025omnipaint} reconceptualizes object removal and insertion as interdependent tasks and introduces a progressive training pipeline. Erase Diffusion~\cite{Liu2025EraseDiffusion} establishes innovative diffusion pathways that facilitate a gradual removal of objects, allowing the model to better understand the erasure intent. While most research focuses on designing better inpainting models via supervised fine-tuning (SFT), our work takes a different approach and analyze a fundamental problem: the tendency of inpainting models to generate unwanted objects within masked regions stems primarily from their over-reliance on surrounding context. Based on this insight, we introduce a spatial-matting trajectory refinement method to develop more effective sampling and training strategies based on GRPO framework, specifically designed to mitigate this issue.

\begin{figure*}[t]
    \centering
    \includegraphics[width=\linewidth]{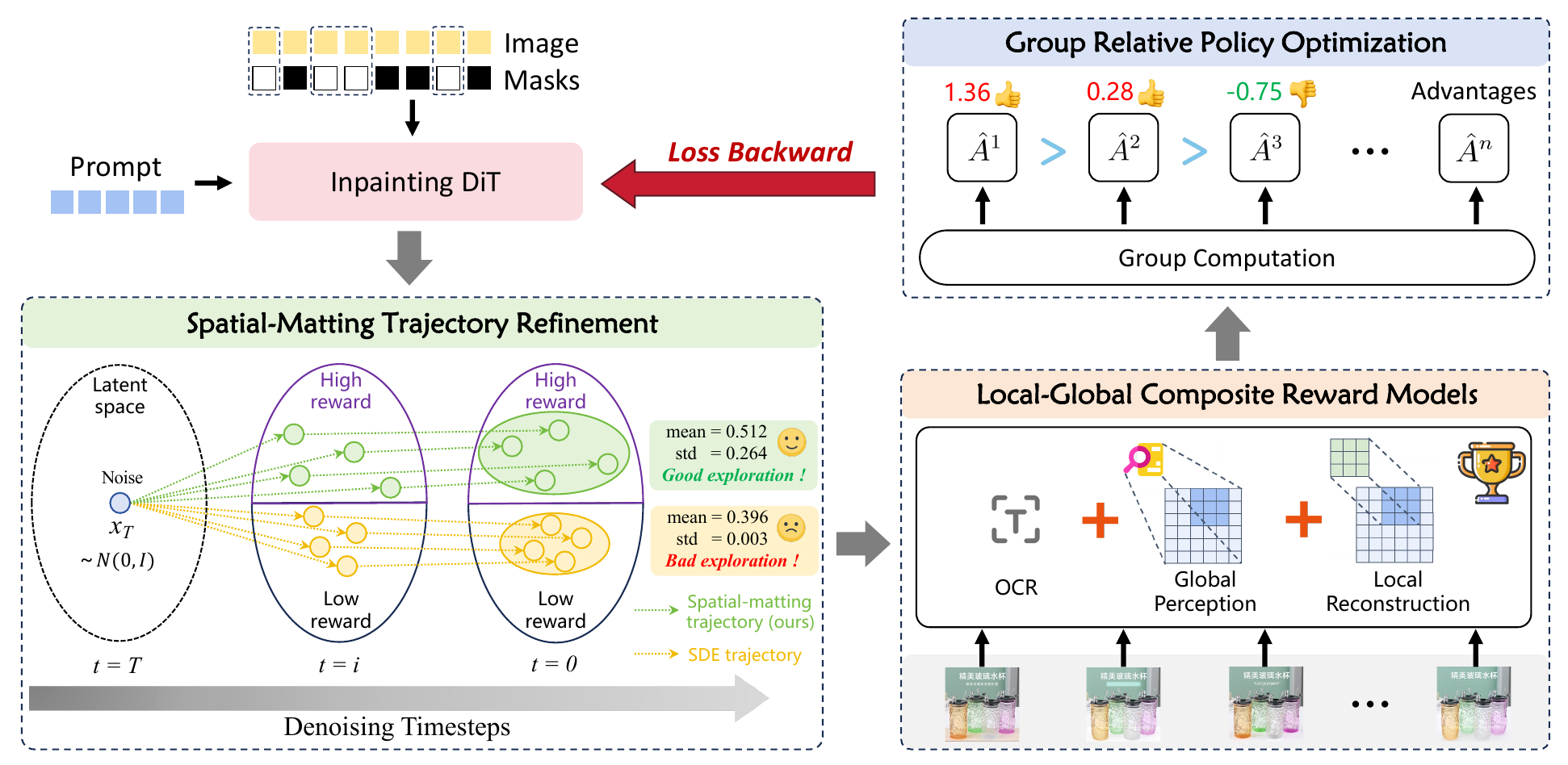}
    \caption{Overview of RePainter. We propose a novel reinforcement learning framework that integrates spatial-matting trajectory refinement with GRPO. The spatial-matting module modulates attention mechanisms to optimize the sampling trajectory during denoising, expanding the exploration space and guiding the generation of higher-reward samples. These samples are then evaluated by our local-global composite reward models, which jointly assesses global structural consistency, local pixel accuracy, and semantic validity. Rewards from these trajectories feed the GRPO loss, enabling online policy updates that align the model with e-commerce-specific visual preferences.}
    \label{fig:pipeline}
    \Description{Figure 3}
\end{figure*}

\section{Preliminary: Flow-based GRPO}
\label{Preliminary: Flow-based GRPO}
In this section, we present the core idea of GRPO applied to flow matching models. We first revisit how flow-based GRPO converts the deterministic ODE sampler into a SDE sampler with the same marginal distribution, which satisfies GRPO’s stochastic exploration requirements. Then, we present the algorithm of flow-based GRPO.

\noindent \textbf{Flow Matching.}
Let $\mathbf{z}_0$ be a data sample from the true distribution and $\mathbf{z}_1$ a noise sample. Rectified flow~\cite{Liu2022FlowStraight} defines intermediate samples as:
\begin{equation}
\mathbf{z}_t = (1-t)\mathbf{z}_0 + t\mathbf{z}_1, \quad t \in [0,1],
\end{equation}
and trains a velocity field $v_\theta(x_t,t)$ via flow matching~\cite{Lipman2022FlowMatching} objective:
\begin{equation}
\mathcal{L}_{\text{FM}}(\theta) = \mathbb{E}_{t,\mathbf{z}_0,\mathbf{z}_1} \big[\|v - v_\theta(\mathbf{z}_t,t)\|_2^2 \big], \quad v = \mathbf{z}_1 - \mathbf{z}_0.
\end{equation}

Beyond training, the iterative denoising process at inference time can be naturally formalized as a Markov Decision Process~\cite{Black2023RL4Diffusion}. At each step $t$, the state is $\mathbf{s}_t=(\mathbf{c},t,\mathbf{z}_t)$, where $c$ denotes the prompt, and $\pi(\mathbf{a}_t\mid\mathbf{s}_t)=p(\mathbf{z}_{t-1}\mid\mathbf{z}_t, \mathbf{c})$ is the probability selecting action $\mathbf{a}_t$ from $z_t$ to $z_{t-1}$. The transition is deterministic, \textit{i.e.}, $s_{t+1}=(\mathbf{c}, t-1, \mathbf{z}_{t-1})$. A reward is only provided at the final step: $r(\mathbf{z}_0,\mathbf{c})$ if $t=0$, and zero otherwise.


\noindent \textbf{Convert ODE to SDE.}
Since GRPO~\cite{Guo2025DeepSeekR1,Shao2024DeepSeekMath} requires stochastic exploration through multiple trajectory samples, where policy updates depend on the trajectory probability distribution and their associated reward signals, DanceGRPO~\cite{Xue2025DanceGRPO} unify the sampling processes of the diffusion model and rectified flows into the form of SDE. For the diffusion model, as demonstrated in~\cite{Song2019GradEstimation,Song2020ScoreSDE}, the forward SDE is given by:
$\mathrm{d}\mathbf{z}_t=f_t\mathbf{z}_t\mathrm{d}t+g_t\mathrm{d}\mathbf{w}$. The corresponding reverse SDE can be expressed as:
\begin{equation}
    \mathrm{d}\mathbf{z}_t=\left(f_t\mathbf{z}_t-\frac{1+\varepsilon_t^2}{2}g_t^2\nabla\log p_t(\mathbf{z}_\mathbf{t})\right)\mathrm{d}t+\varepsilon_tg_t\mathrm{d}\mathbf{w},
\label{sde:ddpm}
\end{equation}
where $\mathrm{d} \mathbf{w}$ is a Brownian motion, and $\varepsilon_t$ introduces the stochasticity during sampling. Similarly, the forward ODE of rectified flow is: $\mathrm{d}\mathbf{z}_t=\mathbf{u}_t\mathrm{d}t$. The generative process reverses the ODE in time. However, this deterministic formulation cannot provide the stochastic exploration required for GRPO. Drawing on insights from~\cite{Albergo2022StochasticInterpolants,Albergo2023StochasticInterpolantsFramework}, an SDE case during the reverse process can be defined as follows:
\begin{equation}
    \mathrm{d}\mathbf{z}_t=(\mathbf{u}_t-\frac{1}{2}\varepsilon_t^2\nabla\log p_t(\mathbf{z}_t))\mathrm{d}t+\varepsilon_t\mathrm{d}\mathbf{w},
\label{sde:rf}
\end{equation}
where $\varepsilon_t$ also introduces the stochasticity during sampling. Given a normal distribution $p_t(\mathbf{z}_t) = \mathcal{N}(\mathbf{z}_t\mid\alpha_t\mathbf{x},\sigma_t^2 I)$, the score function is derived as $\nabla\log p_t(\mathbf{z}_t) = -(\mathbf{z}_t - \alpha_t\mathbf{x})/\sigma_t^2$. This expression can be substituted into the above two SDEs to obtain the policy $\pi(\mathbf{a}_t\mid\mathbf{s}_t)$.

\noindent\textbf{GRPO on Flow Matching.}
GRPO~\cite{Guo2025DeepSeekR1,Shao2024DeepSeekMath} introduces a group-relative advantage to
stabilize policy updates. Given a prompt $\mathbf{c}$, generative models will sample a group of outputs $\{ \mathbf{o}_1, \mathbf{o}_2, ..., \mathbf{o}_G \}$ from the model $\pi_{\theta_{old}}$, and optimize the policy model $\pi_{\theta}$ by maximizing the following objective function:
\begin{equation}
\begin{aligned}
\mathcal{J}(\theta) & = \mathbb{E}_{\substack{\{\mathbf{o}_i\}_{i=1}^G \sim \pi_{\theta_{\text{old}}}(\cdot|\mathbf{c}), \
\mathbf{a}_{t,i} \sim \pi_{\theta_{\text{old}}}(\cdot|\mathbf{s}_{t,i})}} \\
& \bigg[ \frac{1}{G} \sum_{i=1}^G \frac{1}{T} \sum_{t=1}^T \min\bigg( \mathbf{\rho}_{t,i} A_i, \text{clip}\big( \mathbf{\rho}_{t,i}, 1-\epsilon, 1+\epsilon \big) A_i \bigg) \bigg],
\end{aligned}
\label{eq:grpoloss}
\end{equation}

where $\mathbf{\rho}_{t,i} = \frac{\pi_{\theta}(\mathbf{a}_{t,i}|\mathbf{s}_{t,i})}{\pi_{\theta_{old}}(\mathbf{a}_{t,i}|\mathbf{s}_{t,i})} $, and $\pi_{\theta}(\mathbf{a}_{t,i}|\mathbf{s}_{t,i})$ is the policy function is MDP for output $\mathbf{o}_i$ at time step $t$, $\epsilon$ is a hyper-parameter, and $A_i$ is the advantage function, computed using a group of rewards $\{ r_1, r_2,...,r_G \}$ corresponding to the outputs within each group:
\begin{equation}
    A_i=\frac{r_i-\mathrm{mean}(\{r_1,r_2,\cdots,r_G\})}{\mathrm{std}(\{r_1,r_2,\cdots,r_G\})}.
\label{eq:adv}
\end{equation}
Due to reward sparsity in practice, flow-based GRPO~\cite{liu2025flowgrpo,Xue2025DanceGRPO} apply the same reward signal across all timesteps during optimization. Notably, while traditional GRPO formulations employ KL-regularization to prevent reward over-optimization, DanceGRPO~\cite{Xue2025DanceGRPO} empirically observe minimal performance differences when omitting this component and disable this loss.

\section{Methodology}

In this section, we present the details of our proposed RePainter, as illustrated in Figure~\ref{fig:pipeline}. We first elaborate on our sampling refinement strategy based on spatial-matting, followed by the design of local-global composite reward models. Finally, we introduce the EcomPaint-100K dataset and the EcomPaint-Bench.

\subsection{Spatial-matting Trajectory Refinement}

\noindent\textbf{Sparse Rewards and Insufficient Exploration.}
Despite recent progress~\cite{liu2025flowgrpo,Xue2025DanceGRPO} in introducing stochasticity through converting ODEs to SDEs in flow-based models, significant challenges persist during training. Within each group, sample diversity is predominantly governed by the randomness inherent in the SDE process or minor variations in initialization noise. Such restricted variation often results in nearly identical reward signals across samples. This problem becomes particularly pronounced in tasks with sparse reward signals or high complexity, where the model finds it difficult to discover high-reward outputs within a limited exploration space. For instance, when all generated images are invalid and receive equal reward, the group advantage estimate reduces to zero, leading to vanishing policy gradients. As a result, policy updates become ineffective, convergence slows significantly, and overall learning efficiency is severely compromised.

\noindent\textbf{Key Insight.}
Unlike common mitigation strategies in large language models (LLMs), such as dynamic sampling~\cite{Yu2025DAPO} or curriculum learning~\cite{wan2025qwenlongl1,Team2025KimiK1.5}, our core idea is to expand the exploration space at the model level, enabling the sampling process to access a broader and more reward-promoting exploration space. In object removal task, pixels outside the mask serve as “visual prompts” that guide inpainting within the masked area. We observe that in e-commerce images, unwanted insertions, especially text, often occur when the masked region overly references irrelevant foreground elements such as price tags or promotional text. We believe that the generation of the mask region should rely more on background context rather than foreground distractions. The above analysis motivates us to intervene in the generation process and propose a sampling trajectory optimization method using spatial matting. By adjusting the attention layers to calibrate the sampling path, our approach enables the masked region to rely more on background context and ignore distracting areas, which improves the generation of coherent content and suppress the generation of inconsistent artifacts or unexpected objects.

\begin{figure}[t]
    \centering
    \includegraphics[width=\linewidth]{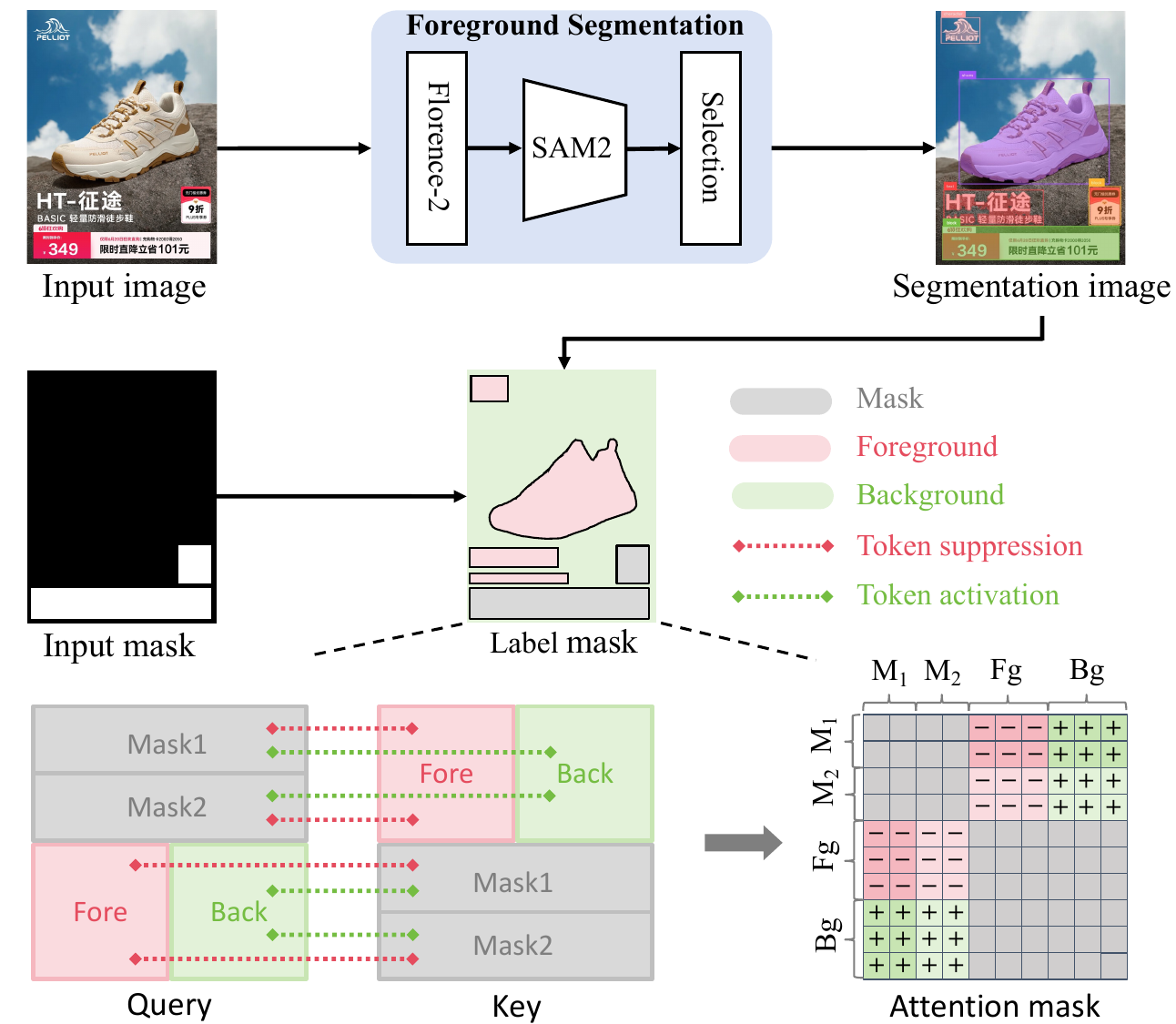}
    \caption{We first apply panoptic segmentation to the image to identify foreground (negative) and background (positive) regions. The spatial-matting strategy aims to make the masked area’s generation more attentive to the background context, while suppressing interference from distracting foreground objects (e.g., price tags or text), thereby reducing the generation of unwanted objects.}
    \label{fig:attention}
\end{figure}

As depicted in Figure~\ref{fig:attention}, we first perform panoptic segmentation on the reference image using SAM2~\cite{Ravi2024SAM2} and Florence2~\cite{Xiao2024Florence2} models, dividing it into three semantic regions: foreground, background, and the mask region to be inpainted. The mask attention operation in MMDiT~\cite{BlackForestLabsFlux2024,Esser2024Scaling} can be expressed as:
\begin{equation}
A' = \text{softmax} \left( \frac{QK^T + M}{\sqrt{d}} \right),
\end{equation}
\begin{equation}
M = W^{pos} \odot M^{pos} + W^{neg} \odot M^{neg},
\end{equation}
where $M \in \mathbb{R}^{1 \times N^2}$ represents the flattened mask, $N$ denotes the total token length in latent space, including both image and text tokens. Based on the segmentation results, these tokens are categorized into four parts: foreground ($r^{fg}$), background ($r^{bg}$), mask region ($r^{m}$), and text ($r^{t}$). To guide the masked region generation to rely more on background references rather than foreground objects, we increase the attention scores between the masked region and background areas while suppressing those toward foreground objects. Specifically, we design semantically aware $M^{pos}$ and $M^{neg}$ based on panoptic segmentation results, where $M^{pos}$ represents positive regions (i.e., background semantics) that should receive stronger attention, while $M^{neg}$ denotes negative regions share semantic similarity with the target objects to be removed. Correspondingly, for each query pixel $i$ and key pixel $j$ in the attention maps, $M^{pos}$ and $M^{neg}$ are defined as follows:
\begin{equation}
M^{pos}_{i,j} = 
\begin{cases}
1, & \text{if } (i, j) \in \{(r^{mask},r^{bg}), (r^{bg},r^{mask})\}, \\
0, & \text{otherwise},
\end{cases}
\end{equation}
\begin{equation}
M^{neg}_{i,j} = 
\begin{cases}
1, & \text{if } (i, j) \in \{(r^{mask},r^{fg}), (r^{fg},r^{mask})\}, \\
0, & \text{otherwise}.
\end{cases}
\end{equation}

Since our method alters the original denoising process, it may potentially compromise the image quality of the pre-trained model. To mitigate this risk, we modulate the weight values $W^{pos}$ and $W^{neg}$ according to the range of original attention scores inspired by~\cite{Li2024MagicEraser,Kim2023DenseDiffusion}. We calculate the following matrices that identify each query’s maximum and minimum values, ensuring the modulated values stay close to the original range. Therefore, the adjustment is proportional to the difference between the original values and either the maximum value (for positive pairs) or the minimum value (for negative pairs):
\begin{align}
W^{pos} &= \max(QK^{\top}) - QK^{\top}, \\
W^{neg} &= \min(QK^{\top}) - QK^{\top}.
\end{align}

Notably, our method optimizes the sampling path via spatial attention operations and is completely training-free. Consequently, it can outperform baseline models even in the absence of any additional training, as detailed in Section~\ref{sec:ablation}. 
Nevertheless, during the roll-out phase of GRPO, we apply spatial-matting strategy to group samples with a probability of $\lambda$, which is empirically set to 0.25. This probabilistic selection is designed to promote greater reward diversity and variability during the early stages of training. 
Moreover, this strategy can be generalized into other application domains as long as the negative and positive regions are well-defined.
\subsection{Local-Global Composite Reward Models}

In GRPO-based image inpainting, it is essential to design a reward function that captures multiple dimensions of image quality in order to effectively guide the policy network. Relying on a single reward often leads to suboptimal results, such as blurred textures or inconsistent artifacts, as shown in Figure~\ref{fig:reward_ablation}. To this end, we propose a composite reward framework that systematically combines global structural guidance with local region-specific incentives. 

\noindent\textbf{Global Structural Reward.}
To assess the overall structural coherence between the generated image and ground truth, we introduce a global structural reward, which evaluates the statistical similarity between corresponding local image patches by comparing their mean-centered patch vectors using cosine similarity. For each pair of corresponding local windows $W_x$ and $W_y$ from the generated image $X$ and ground truth image $Y$, we compute their mean-centered versions $\tilde{\mathbf{x}} = W_x - \mu_x$ and $\tilde{\mathbf{y}} = W_y - \mu_y$, where $\mu_x$ and $\mu_y$ represent the mean pixel values of $W_x$ and $W_y$ respectively. We then calculate variances $\sigma_x^2 = \|\tilde{\mathbf{x}}\|_2^2$, $\sigma_y^2 = \|\tilde{\mathbf{y}}\|_2^2$ and covariance $\sigma_{xy} = \langle\tilde{\mathbf{x}}, \tilde{\mathbf{y}}\rangle$. The local consistency score and global reward are calculated as:
\begin{equation}
S(W_x, W_y) = \frac{ \sigma_{xy} + k }{ \sqrt{\sigma_x^2 \sigma_y^2} + k }, \quad R^{\text{global}} = \frac{1}{N} \sum_{i=1}^{N} S(W_x^{(i)}, W_y^{(i)}),
\end{equation}
where $k$ is a small constant for numerical stability, and $N$ represents the total number of local window pairs from the images. This formulation represents the cosine similarity between mean-centered vectors, producing values in $[-1, 1]$ where 1 indicates perfect match. By evaluating local statistical similarity, $R^{\text{global}}$ provides perceptually-aligned structural guidance for policy optimization.

\begin{algorithm}[t]
\caption{RePainter Training Algorithm}
\label{RePainter_algorithm}
\renewcommand{\algorithmicrequire}{\textbf{Input:}}
\renewcommand{\algorithmicensure}{\textbf{Output:}}
\begin{algorithmic}[1]
\REQUIRE
    Policy model $\pi_\theta$, reward models $\{R_k\}_{k=1}^K$, image-mask pair dataset $\mathcal{D}$, timestep selection ratio $\tau$, total sampling steps $T$.
\ENSURE
    Optimized policy model $\pi_\theta$
\FOR{training iteration $m = 1$ to $M$}
    \STATE Sample batch $\mathcal{D}_b \sim \mathcal{D}$ 
    \STATE Update old policy: $\pi_{\theta_{\text{old}}} \gets \pi_\theta$
    \FOR{each image-mask pair $\mathbf{c} \in \mathcal{D}_b$}
        \STATE Generate $G$ samples: $\{\mathbf{o}_i\}_{i=1}^G \sim \pi_{\theta_{\text{old}}}(\cdot|\mathbf{c})$ with the same random initialization noise
        \STATE Compute rewards $\{r_i^k\}_{i=1}^G$ using $R_k$
        \FOR{each sample $i = 1$ to $G$}
            \STATE Calculate multi-reward advantage: $A_i \gets \sum_{k=1}^K \frac{r_i^k - \mu^k}{\sigma^k}$ 
        \ENDFOR
        \STATE Subsample $\lceil\tau T\rceil$ timesteps $\mathcal{T}_{\text{sub}} \subset \{1..T\}$
        \FOR{each timestep $t \in \mathcal{T}_{\text{sub}}$}
            \STATE Update policy via gradient ascent: $\theta \gets \theta + \eta \nabla_\theta \mathcal{J}$
        \ENDFOR
    \ENDFOR
\ENDFOR
\end{algorithmic}
\end{algorithm}

\noindent\textbf{Local Reconstruction Reward.}
This reward component guides the model to achieve high-fidelity pixel-level reconstruction within the masked region $M$. It enforces precise alignment between the generated content and the ground truth through a normalized error metric that is robust to intensity variations across different images. The reward is computed as:
\begin{equation}
R^{\text{local}} = 1 - \frac{ \| M \odot (X - Y) \|_F^2 }{ \| M \odot Y \|_F^2 + \epsilon },
\end{equation}
where $X$ is the model output image, $Y$ is the ground truth image, $M$ is a binary mask (1 for the inpainted region, 0 otherwise), $\odot$ denotes element-wise multiplication, $\| \cdot \|_F$ is the Frobenius norm (square root of the sum of squared matrix elements), and $\epsilon$ is a small constant ($10^{-8}$) to prevent division by zero. $R^{\text{local}}$ normalizes the error using the pixel intensity values of the ground truth image in the target region, effectively eliminating scale variations across different image contents and providing stable and reliable reconstruction quality signals for model training.

\noindent\textbf{Semantic OCR Reward.}
To ensure semantic coherence and prevent the generation of nonsensical texts in masked regions, we introduce a semantic OCR reward with the following formulation:
\begin{equation}
R_{i}^{\text{ocr}} = 
\begin{cases} 
1 & \text{if } \text{OCR}(X^{i}_{m}) = \emptyset \\
0 & \text{otherwise}
\end{cases}, \quad R^{\text{ocr}} = \frac{1}{N} \sum_{i=1}^{N} R_{i}^{\text{ocr}},
\end{equation}
where $X^{i}_{m}$ represents the $i$-th masked region, $\text{OCR}(\cdot)$ is a pre-trained OCR model~\cite{Cui2025PaddleOCR3}, $\emptyset$ indicates no text detected, and $N$ is the total number of masked regions. This formulation provides a clear reward signal that encourages the model to avoid generating recognizable but nonsensical text content while permitting non-textual content generation, making it essential for e-commerce applications where clean, professional product images are required.

\noindent\textbf{Integrated Reward Framework.}
Rather than directly combining rewards, we aggregate advantage functions, as different reward models often operate on different scales. The overall advantage is defined as follows:
\begin{equation}
A_i = \sum_{k=1}^K \frac{r_i^k - \mu^k}{\sigma^k},
\end{equation}
where $A_i$ is the composite advantage score for the $i$-th sample, $K$ denotes the total number of reward models, $r_i^k$ indicates the reward score of the $i$-th sample for the $k$-th reward model, $\mu^k$ and $\sigma^k$ are the mean and standard deviation across the current batch. This formulation enables balanced and comparable aggregation of diverse reward signals, providing a unified advantage estimate for policy optimization. In summary, our composite reward framework provides dense and precise optimization signals from global structure, local reconstruction, and semantic validity, collectively leading to visually seamless and semantically coherent image inpainting results. The detailed algorithm of our proposed RePainter can be found in Algorithm~\ref{RePainter_algorithm}.

\section{EcomPaint Dataset and Benchmark}
 
To address the challenges in e-commerce object removal, we design a systematic pipeline for high-quality dataset construction. This pipeline yields EcomPaint-100K, where each sample comprises a commodity image, a corresponding mask, and an erased clean image, thus supporting robust object removal research in e-commerce scenarios. Additionally, we randomly select 1,000 samples to form EcomPaint-Bench as a standardized evaluation set. More details can be found in Appendix~\ref{sec:ecomPaint}.

\begin{figure*}[t]
    \centering
    \includegraphics[width=1.0\linewidth]{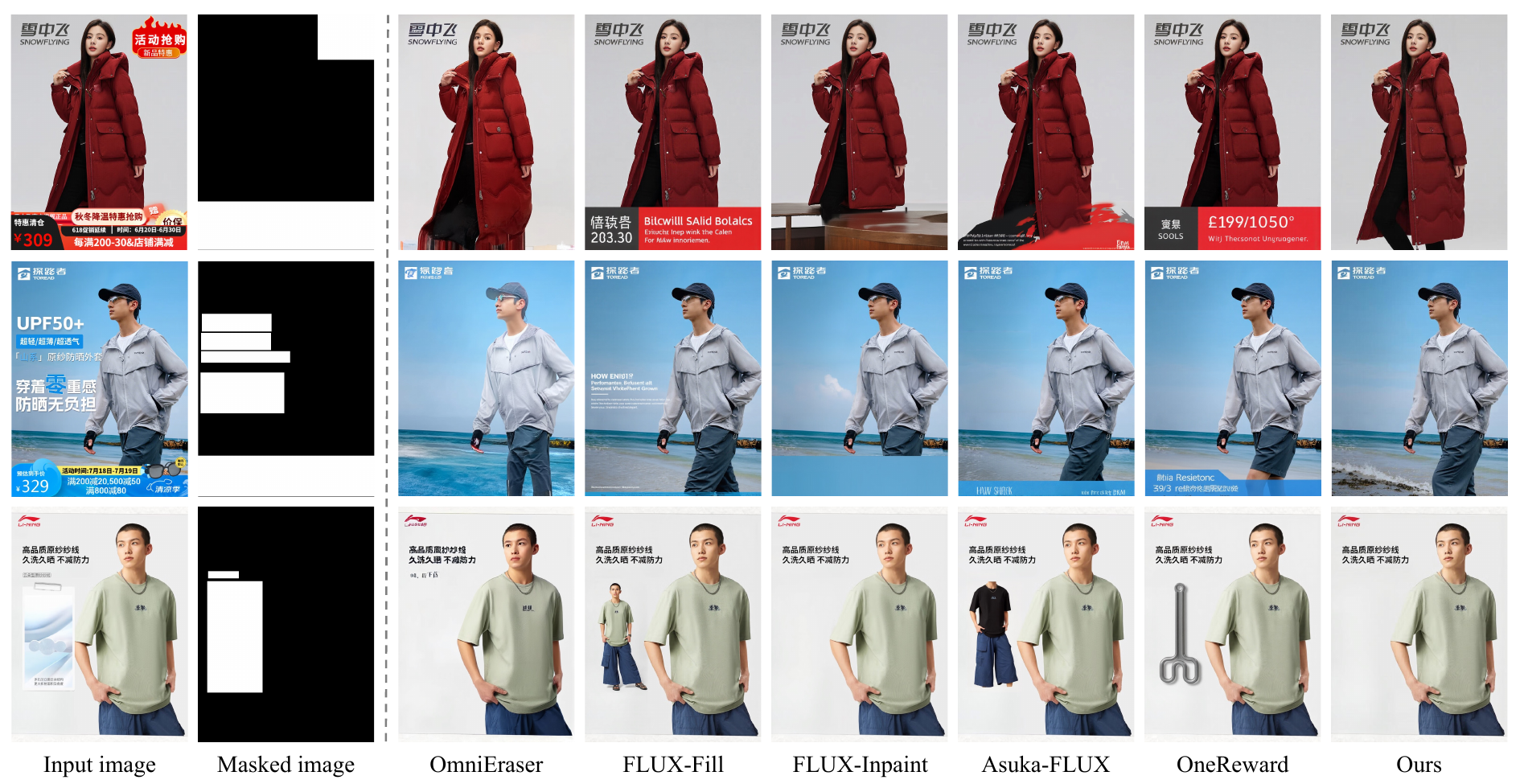}
    \caption{Qualitative results of all comparison methods in challenging scenarios. Our RePainter demonstrates superior capability in unwanted-object-mitigated and structural consistency.}
    \label{fig:compare}
\end{figure*}

\section{Experiments}
\subsection{Experimental Setup}
\noindent\textbf{Implementation Details.}
We use the FLUX.1-Fill~\cite{BlackForestLabsFlux2024} as the base model, which is an advanced inpainting model based on flow matching. We use an empty text prompt for inference and training. The model is fine-tuned for 100 iterations on 8 NVIDIA H100 GPUs with a total batch size of 16 and a resolution of 1024×1024. We use Adam optimizer with the learning rate being 3e-6 during training. 

\noindent\textbf{Comparison Methods.}
We compare our approach with state-of-the-art methods in image inpainting, including OmniEraser~\cite{wei2025omnieraser}, FLUX-Fill~\cite{BlackForestLabsFlux2024}, FLUX-Inpaint~\cite{FluxControlnetInpainting2024}, OneReward~\cite{gong2025onereward} and Asuka~\cite{wang2025towards}. For all models, we use their official weights and default recipes.

\begin{table}[t]
  \centering
  \caption{Quantitative results on the EcomPaint-Bench. The best results are in bold, and the second-best are underlined.}
  \label{tab:quantitative_results}
  \setlength{\tabcolsep}{2pt} 
  \small 
  \begin{tabular}{@{\extracolsep{\fill}}l|ccccccc@{}}
    \toprule
    Methods & FID$\downarrow$ & LPIPS$\downarrow$ & SSIM$\uparrow$ & PSNR$\uparrow$ & P-IDS$\uparrow$ & U-IDS$\uparrow$ & OCR$\uparrow$ \\
    \midrule
    OmniEraser~\cite{wei2025omnieraser} & 3.906 & 0.078 & 0.942 & 28.938 & 0.135 & 0.124 & \textbf{0.964} \\
    FLUX-Fill~\cite{BlackForestLabsFlux2024} & 4.001 & 0.066 & 0.950 & 28.202 & \underline{0.209} & \underline{0.185} & 0.723 \\
    FLUX-Inpaint~\cite{FluxControlnetInpainting2024} & \underline{2.499} & 0.068 & 0.948 & 28.703 & 0.188 & 0.166 & 0.878 \\
    Asuka-FLUX~\cite{wang2025towards} & 3.205 & 0.061 & 0.956 & 28.341 & 0.167 & 0.148 & 0.763 \\
    OneReward~\cite{gong2025onereward} & 4.638 & \underline{0.055} & \underline{0.960} & \underline{29.967} & 0.175 & 0.143 & 0.901 \\
    \textbf{RePainter} & \textbf{1.075} & \textbf{0.033} & \textbf{0.968} & \textbf{36.070} & \textbf{0.259} & \textbf{0.242} & \underline{0.926} \\
    \bottomrule
  \end{tabular}
\end{table}

\noindent\textbf{Evaluation Metrics.}
To assess the quality of the generated images, we report the metrics LPIPS~\cite{Zhang2018lpips} to calculate the patch-level image distances, FID~\cite{Heusel2017fid} to compare the distribution distance between generated images and real images, P-IDS/U-IDS~\cite{Zhao2020CoModganIDS} to measure the human-inspired linear separability, PSNR and SSIM to assess the consistency between the predicted region and corresponding region in the ground truth. We also report the OCR metric to detect abnormal text in masked regions, using the same calculation method as our OCR reward model.

\begin{table}[h]
    \centering
    \caption{Quantitative results of user study and GPT-4o evaluations. Our method achieves the best performance.}
    \setlength{\tabcolsep}{10pt}
    \small 
    \begin{tabular}{l|c|c}
    \toprule
    Method & User Study & GPT Evaluation\\
    \midrule
    OmniEraser~\cite{wei2025omnieraser} & 18.6\% & 8.3\%\\
    FLUX-fill~\cite{BlackForestLabsFlux2024} & 56.5\% & 41.1\%\\
    FLUX-Inpaint~\cite{FluxControlnetInpainting2024} & \underline{68.6\%} & 54.2\%\\
    Asuka-FLUX~\cite{wang2025towards} & 59.9\% & 48.0\%\\
    OneReward~\cite{gong2025onereward} & 67.4\% & \underline{70.3\%} \\
    \textbf{RePainter} & \textbf{85.2\%} & \textbf{73.9\%}\\
    \bottomrule
    \end{tabular}
    \label{tab:user_study}
\end{table}

\subsection{Comparison with Existing Methods}
\noindent\textbf{Quantitative Results.}
Table~\ref{tab:quantitative_results} presents a comprehensive comparison of our method with several state-of-the-art approaches on the EcomPaint-Bench. Our model consistently achieves superior results across all metrics, achieving the highest SSIM, PSNR, P-IDS, and U-IDS while maintaining the lowest FID and LPIPS. These results highlight its ability to remove objects while preserving structural and semantic consistency, effectively suppressing object hallucination. While OmniEraser~\cite{wei2025omnieraser} achieves the highest OCR score, this comes at the cost of inferior results on other image quality metrics, as it fills masked regions with unrealistic content.

\noindent\textbf{Qualitative Results.}
Figure~\ref{fig:compare} illustrates the visualization results of all approaches. The state-of-the-art inpainting algorithms often suffer from unnatural generation. For example, unnatural boundaries and nonsensical text can be observed in the first and second rows, and the inpainting of price tags fails in the third row. OmniEraser~\cite{wei2025omnieraser} sometimes produces blurred results, especially when dealing with large, continuous masks. Flux-Fill~\cite{BlackForestLabsFlux2024}, FLUX-Inpaint~\cite{FluxControlnetInpainting2024}, Asuka~\cite{wang2025towards} and OneReward~\cite{gong2025onereward} frequently exhibit unwanted object insertion and hallucinate unreasonable objects in nearly all illustrated cases. In contrast, our method achieves high-quality inpainting that effectively mitigates unwanted objects and maintains structural consistency.

\noindent\textbf{User Study and GPT-4o Evaluation.}
Due to the lack of effective metrics for the object removal task, the aforementioned metrics may not fully demonstrate the advantages of our method. Therefore, to further validate its effectiveness, we conduct a user study in which participants evaluate whether each image successfully meets the object removal criteria. The overall pass rate is then calculated for each method. As shown in Table~\ref{tab:user_study}, our approach achieves the highest pass rate, which is consistent with the quantitative results and highlights its superior performance. Additionally, we design fair and reasonable prompts and utilize GPT-4o~\cite{OpenAI2024GPT4o} to further assess the object removal capabilities of our method compared to other approaches. The results also show that our method significantly outperforms the alternatives, demonstrating outstanding performance. For more details, please refer to Appendix~\ref{sec:user_study}.

\begin{table*}[t]
    \centering
    \setlength{\tabcolsep}{4pt}
    \small
    \caption{Ablation study on our proposed spatial-matting trajectory refinement and GRPO-training strategy. Blue shows performance gain over the baseline (the first row).}
    \label{tab:ablation}
    \begin{tabular}{cc|ccccccc}
    \toprule
    Spatial-matting & GRPO-training & FID$\downarrow$ & LPIPS$\downarrow$ & SSIM$\uparrow$ & PSNR$\uparrow$ & P-IDS$\uparrow$ & U-IDS$\uparrow$ & OCR$\uparrow$ \\
    \midrule
    \xmark & \xmark & 4.001 & 0.066 & 0.950 & 28.202 & 0.209 & 0.185 & 0.723 \\
    \cmark & \xmark & 2.290\textcolor{blue}{(-1.711)} & 0.054\textcolor{blue}{(-0.012)} & 0.958\textcolor{blue}{(+0.008)} & 31.019\textcolor{blue}{(+2.817)} & 0.226\textcolor{blue}{(+0.017)} & 0.185\textcolor{blue}{(0.000)} & 0.857\textcolor{blue}{(+0.134)} \\
    \xmark & \cmark & 1.263\textcolor{blue}{(-2.738)} & 0.036\textcolor{blue}{(-0.030)} & 0.966\textcolor{blue}{(+0.016)} & 35.010\textcolor{blue}{(+6.808)} & 0.230\textcolor{blue}{(+0.021)} & 0.204\textcolor{blue}{(+0.019)} & 0.909\textcolor{blue}{(+0.186)} \\
    \cmark & \cmark & \textbf{1.075}\textcolor{blue}{(-2.926)} & \textbf{0.033}\textcolor{blue}{(-0.033)} & \textbf{0.968}\textcolor{blue}{(+0.018)} & \textbf{36.070}\textcolor{blue}{(+7.868)} & \textbf{0.259}\textcolor{blue}{(+0.050)} & \textbf{0.242}\textcolor{blue}{(+0.057)} & \textbf{0.926}\textcolor{blue}{(+0.203)} \\
    \bottomrule
    \end{tabular}
\end{table*}

\subsection{Ablation Study}
\label{sec:ablation}

\begin{figure*}[t]
    \centering
    \includegraphics[width=1.0\linewidth]{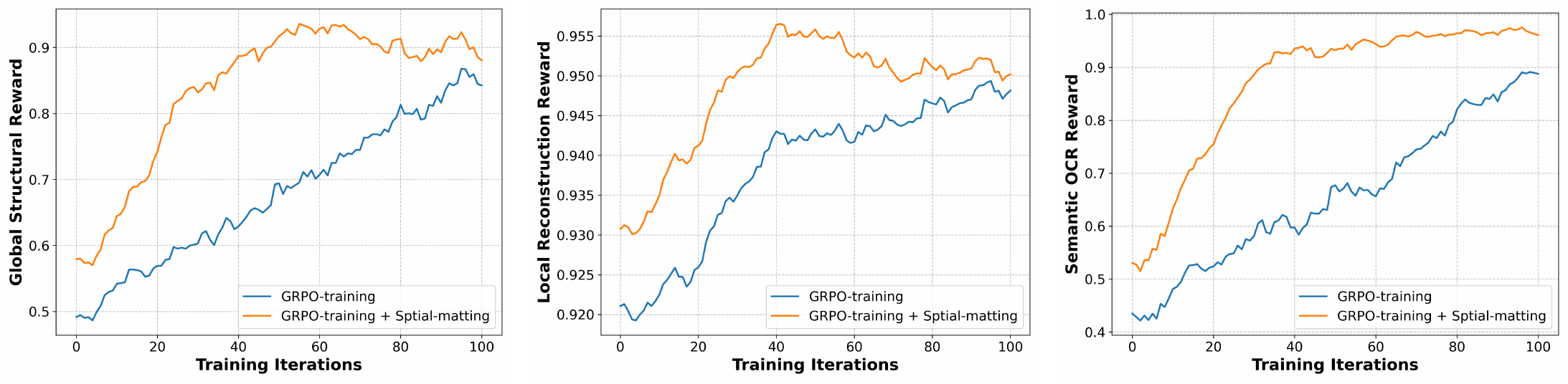}
    \caption{A comparative analysis between standard GRPO-training and our spatial-matting trajectory refinement. We visualize the reward curves of our three proposed reward models. After applying spatial-matting, our model achieves optimal performance across all reward models while requiring fewer training iterations optimized.}
    \label{fig:reward_compare}
    \Description{Figure 3}
\end{figure*}

\noindent\textbf{Effectiveness of Spatial-matting and GRPO-training.}
Table~\ref{tab:ablation} and Figure~\ref{fig:reward_compare} present a comprehensive ablation analysis of the proposed spatial-matting refinement and multi-reward GRPO-training. The results clearly demonstrate that both components independently contribute significant improvements over the baseline across multiple evaluation metrics. Specifically, integrating spatial-matting alone significantly improves the performance for object removal, evidenced by notable gains in FID, PSNR, and OCR metrics. Similarly, GRPO-training independently enhances image quality and semantic coherence, with clear improvements in FID, LPIPS, and perception-related scores.

Most importantly, when both are combined, the model achieves the best overall performance across all metrics, as highlighted by the bolded results in Table~\ref{tab:ablation}. In addition to superior quantitative scores, Figure~\ref{fig:reward_compare} shows that the joint application of spatial-matting and GRPO-training enables the model to converge much faster during training, requiring fewer iterations to reach optimal reward values for all three proposed reward models. The results underscore the complementary advantages of spatial-matting and GRPO-training, which together are essential for stable object removal, high-fidelity image generation, and efficient training.

\begin{figure}[t]
    \centering
    \includegraphics[width=1.0\linewidth]{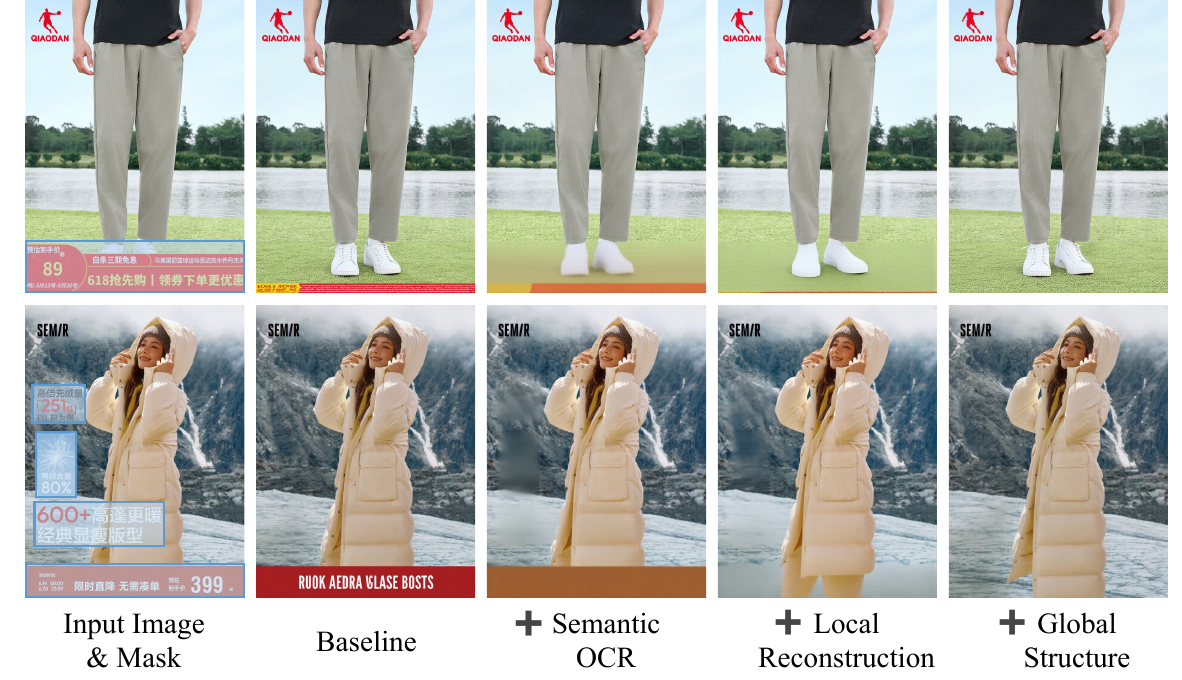}
    \caption{Qualitative ablation comparison of our proposed local-global composite reward models. From left to right, we progressively add each proposed reward model.}
    \label{fig:reward_ablation}
\end{figure}

\noindent\textbf{Effectiveness of composite reward models.}
To further analyze the contribution of each component in our proposed composite reward models, we conduct a qualitative ablation comparison, with results shown in Figure~\ref{fig:reward_ablation}. The baseline model tends to generate undesired text within the masked regions. While training with only the OCR reward successfully eliminates these artifacts, it leads to noticeable color inconsistencies as a result of reward hacking. The addition of the local reconstruction reward mitigates this problem, yet subtle boundary artifacts remain visible. Finally, by incorporating the global structure reward, the model effectively captures structurally coherent guidance and achieves optimal inpainting performance, producing seamless and visually consistent outputs.

\section{Conclusion}
In this paper, we present RePainter, a reinforcement learning framework for e-commerce object removal that integrates spatial-matting trajectory refinement with GRPO. RePainter effectively eliminates unwanted advertising elements while preserving visual coherence and semantic consistency by modulating spatial attention to prioritize background context and suppress distracting foreground references. The proposed composite reward mechanism jointly optimizes global structural consistency, local pixel-level accuracy, and semantic validity, significantly reducing artifacts and preventing reward hacking. To facilitate research in e-commerce-centric inpainting, we contribute EcomPaint-100K, a large-scale, high-quality dataset, along with a standardized benchmarking suite. Extensive experiments demonstrate that RePainter significantly outperforms state-of-the-art methods across both quantitative metrics and human evaluations, setting a new benchmark for robust and high-fidelity object removal in commercial imagery.


\bibliographystyle{ACM-Reference-Format}
\bibliography{sample-base}


%
\appendix

\section{Relevance to the Web}
This work is intrinsically relevant to the Web and E-commerce domains, as it addresses a critical challenge faced by modern e-commerce platforms: visual clutter from excessive advertising elements in product images, which directly impacts user engagement and purchasing behavior online. Our proposed framework, RePainter, enables high-fidelity object removal through reinforcement learning, offering an automated solution to enhance the visual quality of product listings and improve the online shopping experience. By focusing on the unique requirements of e-commerce imagery, our work not only leverages Web artifact (e.g., product images) but also addresses a core Web-related scientific problem: how to automatically generate clean, trustworthy, and appealing visual content in a domain where image quality directly influences user trust and commercial success.

\section{Dataset Construction Pipeline}
\label{sec:ecomPaint}

\begin{figure}[h]
    \centering
    \includegraphics[width=1.0\linewidth]{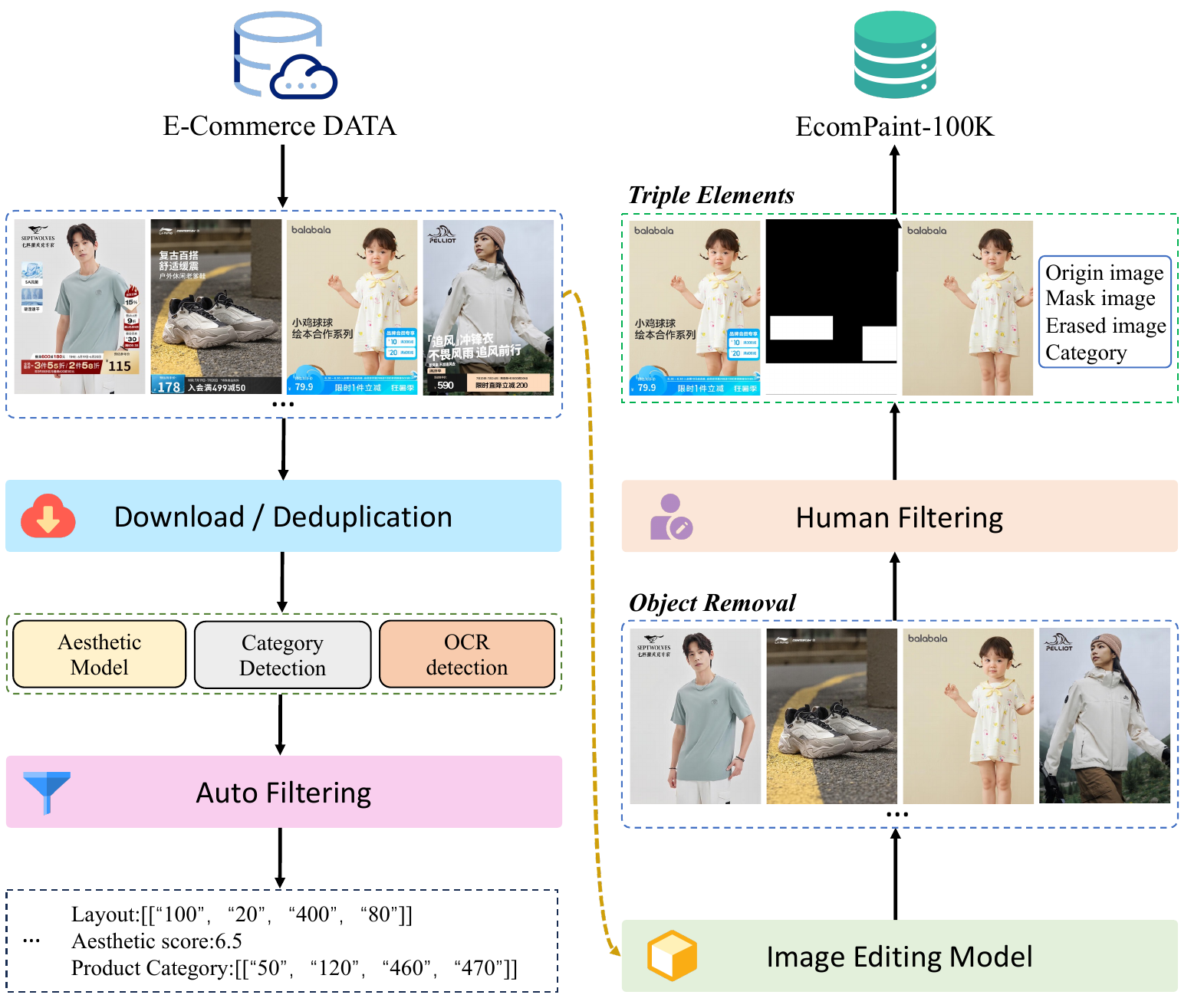}
    \caption{The data construction pipeline of EcomPaint.}
    \label{fig:data_pipeline}
\end{figure}

\begin{figure*}[t]
    \centering
    \includegraphics[width=1.0\linewidth]{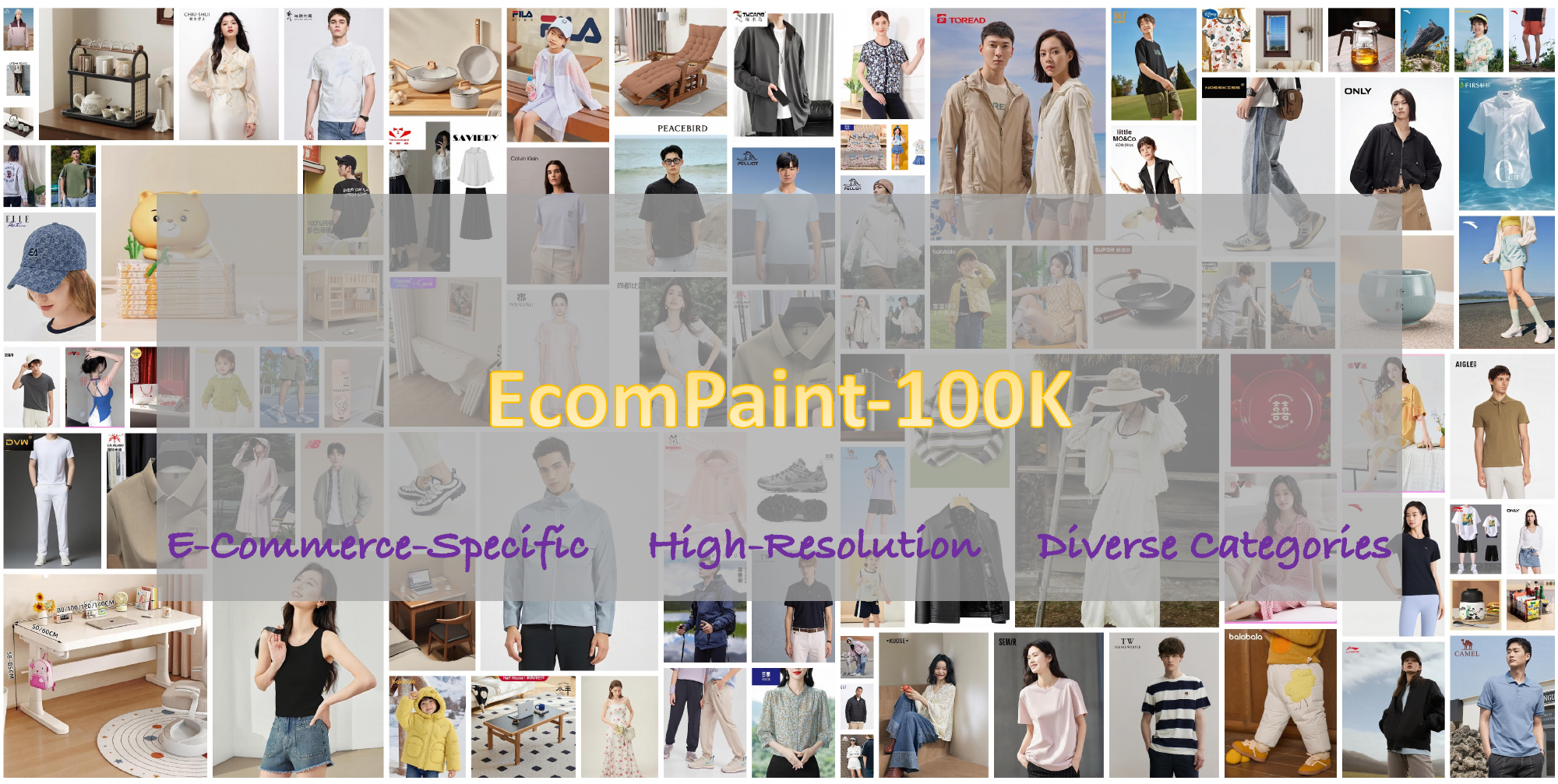}
    \caption{We introduce EcomPaint, a large-scale dataset focused on the e-commerce domain. EcomPaint contains over 100,000 high-resolution image triplets, covering a wide variety of product categories and visual styles. Designed specifically for e-commerce scenarios, this dataset provides a rich foundation for object removal task. EcomPaint aims to accelerate the development of more general and efficient models for e-commerce applications.}
    \label{fig:ecompaint}
\end{figure*}

As shown in Figure~\ref{fig:data_pipeline}, our data construction process begins with collecting a large number of e-commerce images, followed by downloading and deduplication to ensure data quality and uniqueness. Next, we apply automated filtering using aesthetic models, category detection, and OCR detection to select images based on layout, aesthetic score, and product category. After filtering, we use an image editing model to perform object removal and generate the required image variants. We then conduct human filtering to further refine the dataset. Finally, we build the EcomPaint-100K dataset (see Figure~\ref{fig:ecompaint}), where each data sample consists of the original image, mask image, erased image, and category information.

\section{User Study and GPT-4o Evaluation}
\label{sec:user_study}

\begin{figure*}[t]
    \centering
    \includegraphics[width=1.0\linewidth]{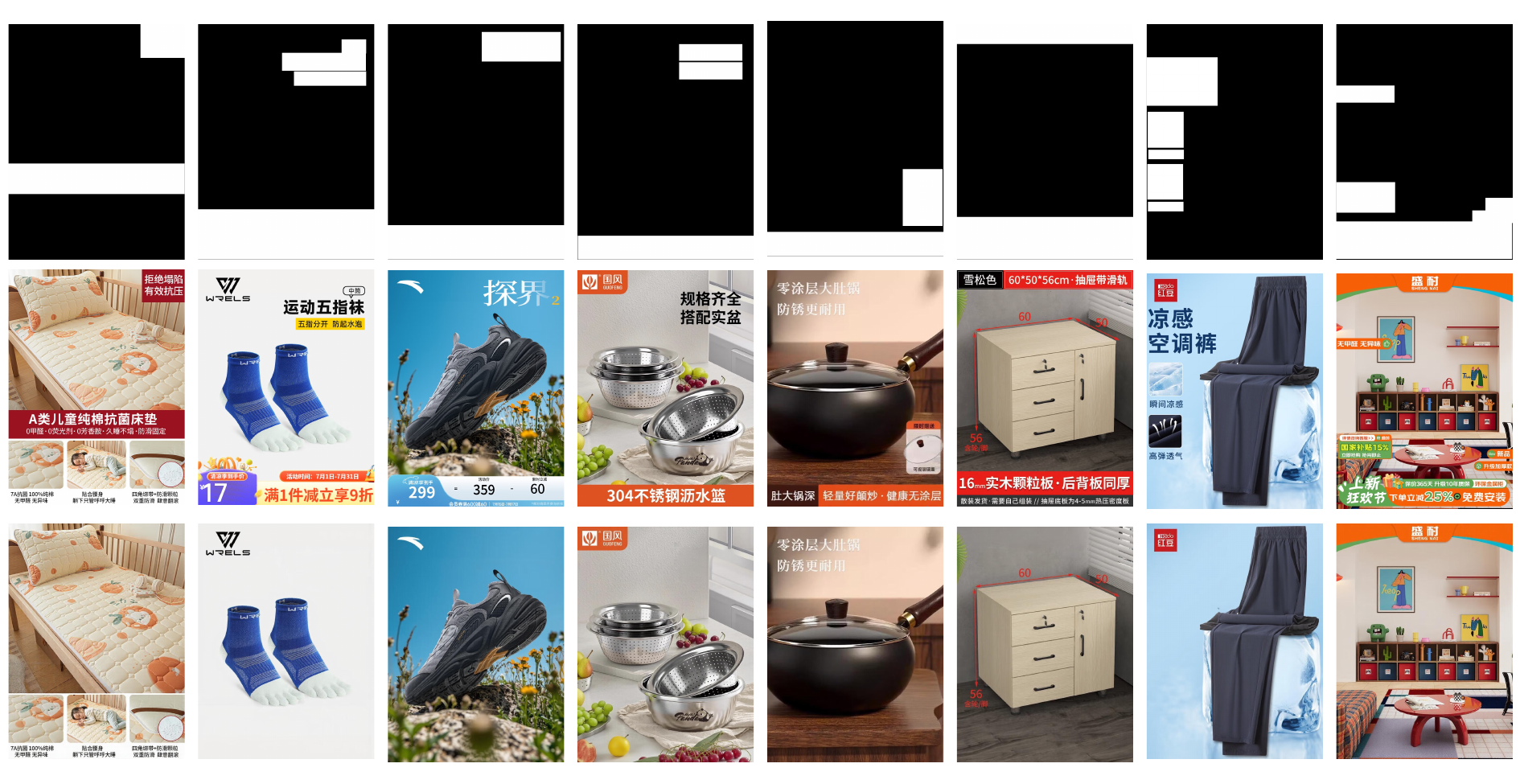}
    \caption{Extensive visual examples of our RePainter applied to diverse product categories.}
    \label{fig:more_vis}
\end{figure*}

In our user study, professional data annotators were recruited to evaluate each inpainted sample based on the following criteria:

\begin{enumerate}
\item The generated content in the erased region is reasonable and well-integrated with the surrounding background.
\item No visible removal traces or logically inconsistent objects/text appear in the erased area.
\item The original product style, details, and clarity of marketing elements (e.g., brand logos, text, patches) are preserved without alteration.
\end{enumerate}
Each evaluation was conducted in randomized order with anonymized samples to ensure fairness. The final object removal performance is reported as the average pass rate across all methods.

For the GPT-4o evaluation, we used the following structured prompt to guide the assessment:
\begin{quote}
You are a professional image quality evaluation expert. Please rigorously assess the effectiveness of image inpainting based on the provided images, which include:
\begin{enumerate}
\item A binary mask image (white indicates the region to be repaired/removed)
\item The original complete image
\item The generated image after object removal
\end{enumerate}

Evaluate based on the following criteria:
\begin{enumerate}
\item \textbf{Removal effectiveness:} Whether the target object is fully removed and the generated content aligns with the surrounding background.
\item \textbf{Visual realism:} Whether the inpainted region appears realistic, without blurriness, artifacts, or unnatural traces.
\end{enumerate}

Provide a binary result as follows:
\begin{itemize}
\item \texttt{1}: Object fully removed and result is realistic/natural.
\item \texttt{0}: Object not fully removed or result is unrealistic.
\end{itemize}

Output format (keep reasoning concise):
\begin{verbatim}
{"score": [0/1], "reasoning": "..."}
\end{verbatim}
\end{quote}
For each method, we performed three independent runs on the evaluation set using different random seeds. The final score is the average pass rate over these runs.

\section{More Visual Examples}

As illustrated in Figure~\ref{fig:more_vis}, we provide a broad range of additional examples demonstrating the performance of our proposed RePainter framework. These diverse visual results showcase the remarkable versatility and effectiveness of our method across a wide spectrum of e-commerce product categories. From electronics and fashion items to household goods and specialty products, RePainter consistently removes undesired elements while generating coherent and structurally consistent content within the masked regions. 



\end{document}